\documentclass[letterpaper, 10 pt, conference]{ieeeconf}  

\usepackage{amsmath}
\usepackage{amssymb}
\usepackage{booktabs}
\usepackage{multirow}
\usepackage{graphicx}
\usepackage{paralist}
\usepackage{wrapfig}
\usepackage{lipsum}
\usepackage{stfloats}
\usepackage{algorithm}
\usepackage{algpseudocode}
\usepackage[caption=false]{subfig}
\usepackage{hyperref}
\hypersetup{hidelinks}

\IEEEoverridecommandlockouts                              

\makeatletter
\def\footnoterule{\relax%
  \kern-5pt
  \hbox to \columnwidth{\vrule width 0.5\columnwidth height 0.4pt\hfill}
  \kern4.6pt}
\makeatother

\title{\LARGE \bf
SEIL: Simulation-augmented Equivariant Imitation Learning
}

\author{Mingxi Jia$^*$, Dian Wang$^*$, Guanang Su, David Klee, Xupeng Zhu, Robin Walters, Robert Platt\\
Khoury College of Computer Sciences\\ Northeastern University\\
\tt\small{\{jia.ming, wang.dian, su.gu, klee.d, zhu.xup, r.walters, r.platt\}@northeastern.edu}
\thanks{*Equal contribution}
}

\newcommand{\SO}{\mathrm{SO}}
\newcommand{\OO}{\mathrm{O}}

\begin{document}

\maketitle
\thispagestyle{empty}
\pagestyle{empty}

\begin{abstract}

In robotic manipulation, acquiring samples is extremely expensive because it often requires interacting with the real world.  Traditional image-level data augmentation has shown the potential to improve sample efficiency in various machine learning tasks. However, image-level data augmentation is insufficient for an imitation learning agent to learn good manipulation policies in a reasonable amount of demonstrations. We propose Simulation-augmented Equivariant Imitation Learning (SEIL), a method that combines a novel data augmentation strategy of supplementing expert trajectories with simulated transitions and an equivariant model that exploits the $\OO(2)$ symmetry in robotic manipulation. Experimental evaluations demonstrate that our method can learn non-trivial manipulation tasks within ten demonstrations and outperforms the baselines with a significant margin.


\end{abstract}

\section{INTRODUCTION}
Policy learning in robotic manipulation is a challenging problem because it is data hungry. When policy learning is modeled using reinforcement learning~\cite{sutton2018reinforcement}, the agent needs to learn from trial and error while interacting with the environment. This process typically requires thousands or millions of time steps, which is extremely expensive even in a simulator. An alternative approach is to model policy learning as imitation learning~\cite{fang2019survey,ibc} by cloning an expert's behavior in a supervised way. Even then, the agent still needs hundreds of episodes of data to learn a simple task like object picking~\cite{zhang2018deep}.

One way to combat the problem of high data requirements is to use data augmentation to improve the sample efficiency, i.e., helping the neural network to learn more general features from a limited amount of data. Recently, data augmentation has led to many successes in both reinforcement learning~\cite{zhong2020random} and imitation learning~\cite{mandi2022towards} because it not only creates extra training samples, but also acts like a regularizer to prevent neural networks from overfitting~\cite{kostrikov2020image}. However, current data augmentation techniques are typically implemented through image transformation~\cite{understandaug} including cropping, blurring, rotating, etc, limiting their potential in 3D robotic manipulation tasks. 

\begin{figure}[t]
\includegraphics[width=\linewidth]{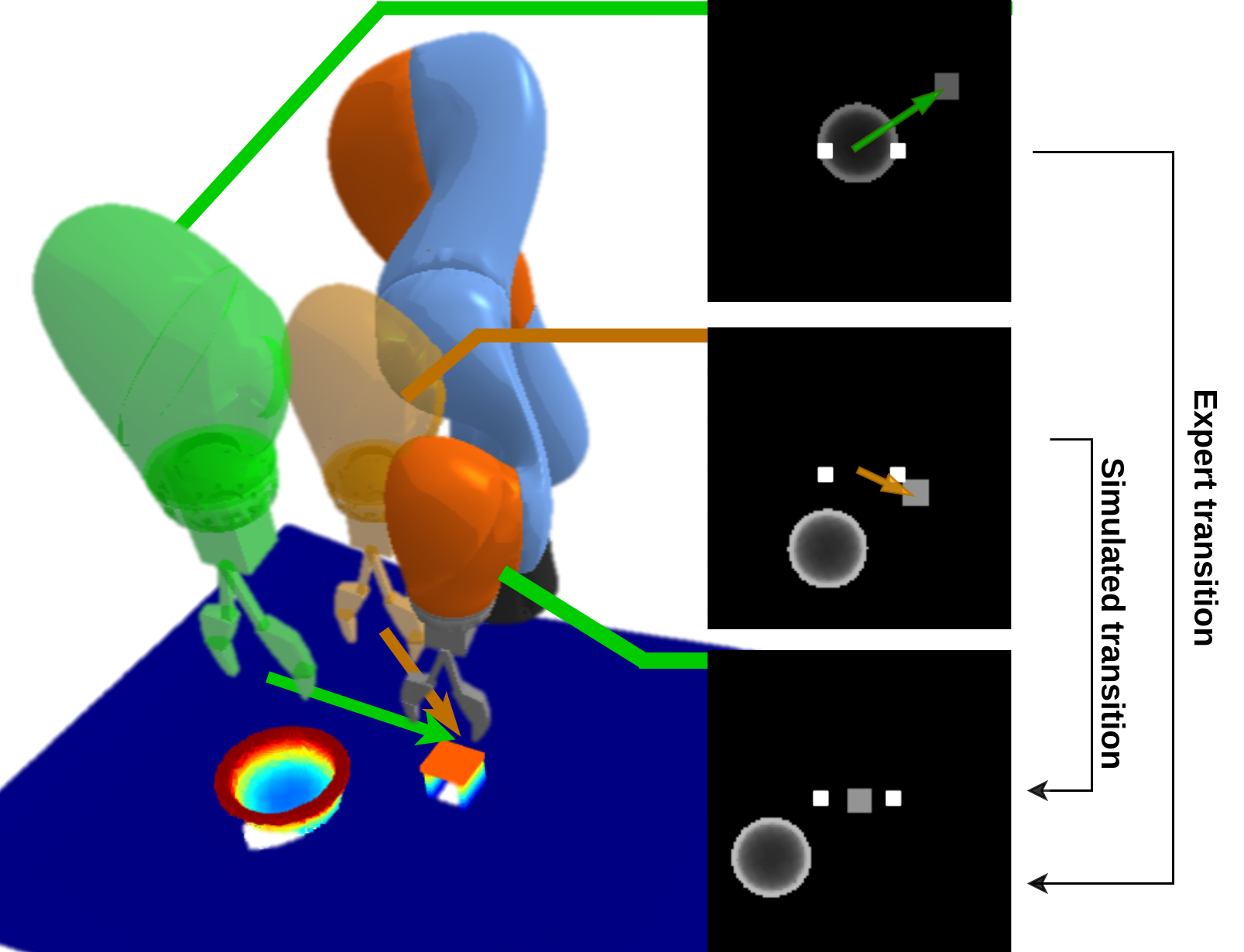}
\caption{\textbf{Overview of Transition Simulation.} Our method generates new expert transitions by projecting the observed point cloud (bottom) to an observation image (right) in a simulated arm pose (orange). Green shows the existing expert transition, and orange shows the simulated transition.}
\label{fig:overview}
\end{figure}

In this work, 
we propose a novel expert data augmentation method that we call \textbf{T}ransition \textbf{S}imulation (TS). Our method simulates new expert state-action pairs within the vicinity of existing expert transitions by projecting observation images from the observed point cloud (Figure~\ref{fig:overview}), resulting in more diverse augmentations compared with standard data augmentation techniques. Transition Simulation can also be combined with image-level data augmentation to further improve the sample efficiency. In addition, we combine Transition Simulation with equivariant policy learning, utilizing the existing $\OO(2)$ (i.e., planar rotation and reflection) symmetry in robotic manipulation to make the policy automatically generalize over different $\OO(2)$ transformed states. Our proposed algorithm, \textbf{S}imulation-augmented \textbf{E}quivariant \textbf{I}mitation \textbf{L}earning (SEIL), achieves few-shot imitation learning using ten or fewer demonstrations in non-trivial manipulation tasks.

Our contributions can be summarized as three-fold. First, we propose a novel data augmentation method, Transition Simulation, that augments the expert data in imitation learning by simulating new expert transitions. Second, we combine Transition Simulation with equivariant policy learning and propose the SEIL method that achieves extremely high sample efficiency in robotic manipulation imitation learning. Last, we evaluate SEIL on a set of challenging manipulation tasks both in simulation and in the real world, and demonstrate a significant improvement compared with competing baselines. Supplementary video and code are available at https://saulbatman.github.io/project/seil/.

\section{RELATED WORK}

\textbf{Imitation learning} or Behavior Cloning (BC) has been widely used by the robotics community. The traditional BC algorithms typically learn an explicit model that maps from the state space to the action space in the framework of supervised learning. Such an approach has been applied to Autonomous driving \cite{bojarski2016end} and robotic manipulation \cite{q-attention,coarse2fine,bchuman,zeng2020transporter,equi_transporter}. Recently, implicit behavior cloning~\cite{ibc} is the first behavioral cloning algorithm to use an energy-based model (EBM), showing impressive results for learning discontinuous behaviors in robotic manipulation. However, implicit models need large amounts of data and careful hyperparameter tuning for their sampling strategy. Compared with prior works, this paper proposes a novel method to improve imitation learning through simulating new expert transitions.

\textbf{Manipulation learning} is a challenging problem because of the high dimensionality and complexity of the robot arm actions. One way to combat that challenge is to use open-loop control with some pre-defined action primitives (e.g., pick, place, push, etc.) to reduce the action space~\cite{zeng2020transporter,asr}. On the other hand, learning in a closed-loop manner~\cite{viereck2020learning,kilinc2022reinforcement,cabi2019scaling,zhan2020framework} is appealing because it allows the agents to behave more freely without being limited by the pre-defined action choices. However, it also makes the policy harder to learn because of the continuous action space and the longer time horizon brought on by the nature of the closed-loop control. This paper solves such a problem though improving the expert data sample efficiency, thus providing more guidance for closed-loop policy learning.

\textbf{Data augmentation} is an indispensable technique in deep learning to improve sample efficiency. Traditionally, data augmentation means image-level augmentation like rotation, reflection~\cite{alexnet}, and translation~\cite{zeng2020transporter}. In particular, random crops~\cite{zhong2020random} have been shown to improve performance in image classification~\cite{alexnet}, reinforcement learning~\cite{rad}, and self-supervised learning~\cite{oord2018representation}. However, image-level augmentation is not generalizable to the 3D space in which the manipulation agent operates. Some recent works perform novel data augmentation in the 3D space. \cite{aug4manipulation} proposes to apply rigid body transformations for states and actions, but requires the ground-truth states of the objects, which makes it highly reliant on high-precision state estimation algorithms in real-world scenarios. \cite{q-attention} augments expert trajectory by connecting intermediate points to keyframes, but defines a keyframe decision function manually, which may possibly lead to inaccurate augmentations. In such cases, RL agents can recover from such failures due to their exploration strategy, but a behavioral cloning agent will suffer from sub-optimality. Compared with the prior works, our method implements data augmentation in the 3D space, does not depend on any state estimators, and ensures the quality of the augmented expert transitions by simulating them near the existing expert transitions. 

\textbf{Data efficiency} is important in terms of the need to avoid labor-heavy and time-consuming data labeling. Besides data augmentation, another approach is to pre-train neural networks with self-supervised learning methods like contrastive learning \cite{chen2020simple} and auto-encoders~\cite{vincent2008extracting}. Such approaches are widely adopted by manipulation works~\cite{li20223d,sermanet2018time,Lee2019}. Recent works in geometric machine learning show that one can improve the data efficiency more directly by enforcing symmetries using equivariant models \cite{e2cnn}, \cite{steerable_cnns}. Several works have applied equivariant models in robotic manipulation and show promising results~\cite{equi_q,so2rl,equi_transporter,zhu2022sample}. 
This works extends the prior works by applying equivariant models in conjunction with Transition Simulation to further improve the sample efficiency.


\section{PROBLEM STATEMENT}
\begin{figure}[t]
\subfloat[PyBullet environment]{\includegraphics[width=0.51\linewidth]{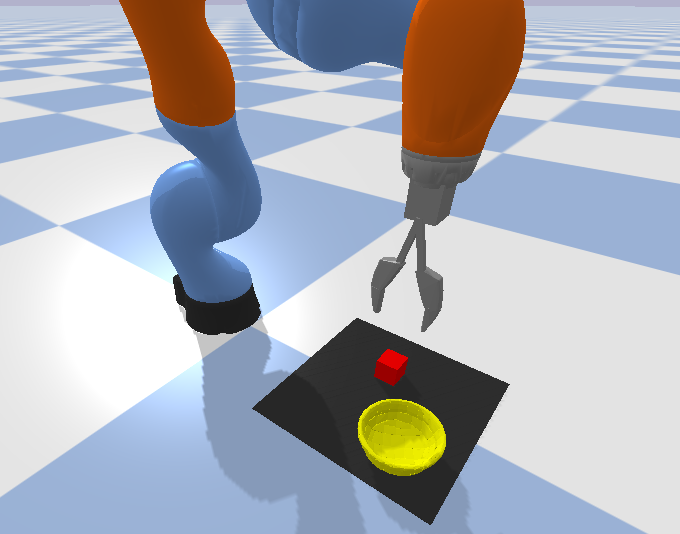}}
\subfloat[Observation in simulation]{\includegraphics[width=0.49\linewidth]{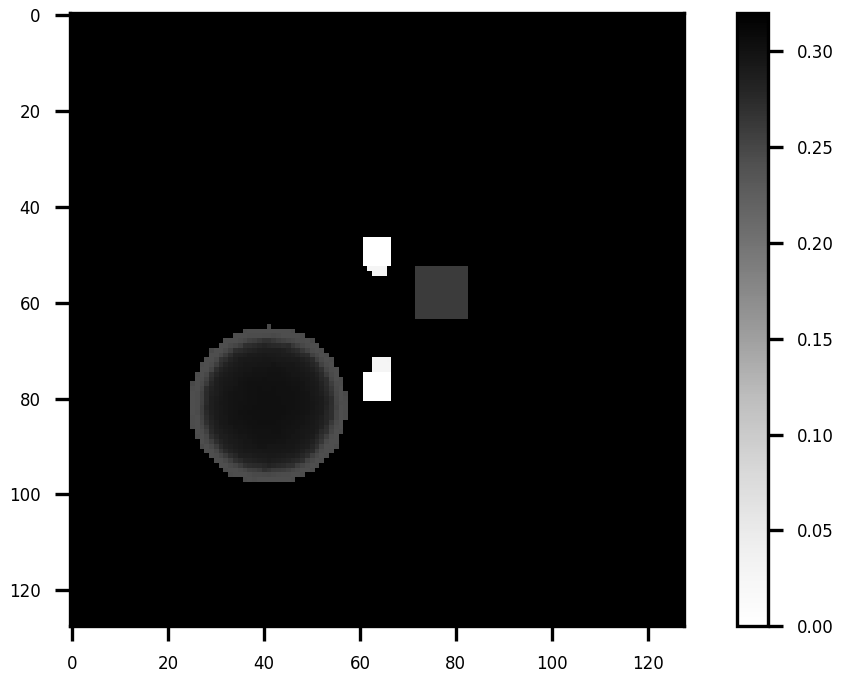}}\\
\caption{\textbf{Environments and observations} (a) Block in Bowl Scene in PyBullet simulation; (b) Depth image observation.}
\label{fig:obs}
\end{figure}
We consider the problem of manipulating (e.g., pick \& place, pushing, etc.) objects in a 3D space in a closed-loop manner. 
This problem can be formulated as learning a neural model mapping a state to the desired delta actions in a 3D workspace $\mathcal{W} \subseteq \mathbb{R}^3$. Let $\mathcal{T} \subseteq \mathcal{W}$ denote the subset of our workspace that is free of objects. 

\textbf{State space}: State is expressed as a tuple $s=(C, p, r)$ where $C\in \mathbb{R}^{n\times 3}$ is a point cloud, $p\in\mathbb{R}^d\subseteq \mathbb{R}^6$ is the $d \leq 6$ degrees of freedom gripper pose, and $r$ is the other robot state expressed arbitrarily. We assume that the robot arm is segmented out from the point cloud $C$. 


\textbf{Action space}: The actions are represented by $a=(\lambda, \Delta p)$ where $\lambda$ is the gripper open width, and $\Delta p$ is the delta movement of the gripper with respect to the current pose $p$. 


\textbf{Five DoF imitation learning with an image observation}: In this work, we focus on gripper poses with $d=4$ degrees of freedom (i.e., $p=(x, y, z, \theta)$ where $\theta$ is the top-down rotation; $\Delta p = (\Delta x, \Delta y, \Delta z, \Delta \theta)$) and learning policies from an image observation. The total number of DoF in the action is five because the agent also needs to control the gripper open width $\lambda$. We define $\Pi: s \mapsto I\in \mathbb{R}^{h\times w}$ as an observation projection function that projects the point cloud $C$ into an orthographic projection depth image $I$ centered at the $(x, y, z)$ position of the gripper. The gripper is located at the center of $I$ with an orientation $\theta$
(Figure~\ref{fig:obs}). Notice that the image $I$ effectively encodes the current gripper pose $p$ 
inside the image such that a planar rotation or reflection of the state will lead to a planar rotation or reflection on the image. Given an expert transition $T = (s, a, s')$, an observation-action pair can be created by $(I=\Pi(s), a)\in B$ where $B$ is the buffer of all observation-action pairs. Let $\pi^*: s\mapsto a$ be the expert policy. Our goal is to learn a neural network $\pi: I\mapsto a$ from $B$ such that $||\pi^*(s) - \pi(I)||_2\to 0$.

\section{APPROACH}

We propose a novel imitation learning algorithm, \textbf{S}imulation-augmented \textbf{E}quivariant \textbf{I}mitation \textbf{L}earning (SEIL). Our method has two main components: a novel augmentation strategy that we call Transition Simulation (TS) and an equivariant model that utilizes the geometric symmetry in the manipulation task. 

\subsection{Transition Simulation}
\label{sec:ts}
\begin{figure}[t]
\centering
\subfloat[Transition simulation on multiple trajectories]{\includegraphics[width=0.8\linewidth]{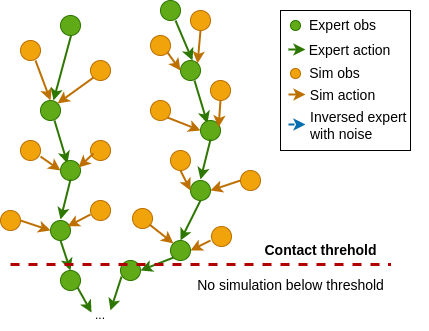}}\\
\centering
\subfloat[Transition simulation on one transition]{\includegraphics[width=0.65\linewidth]{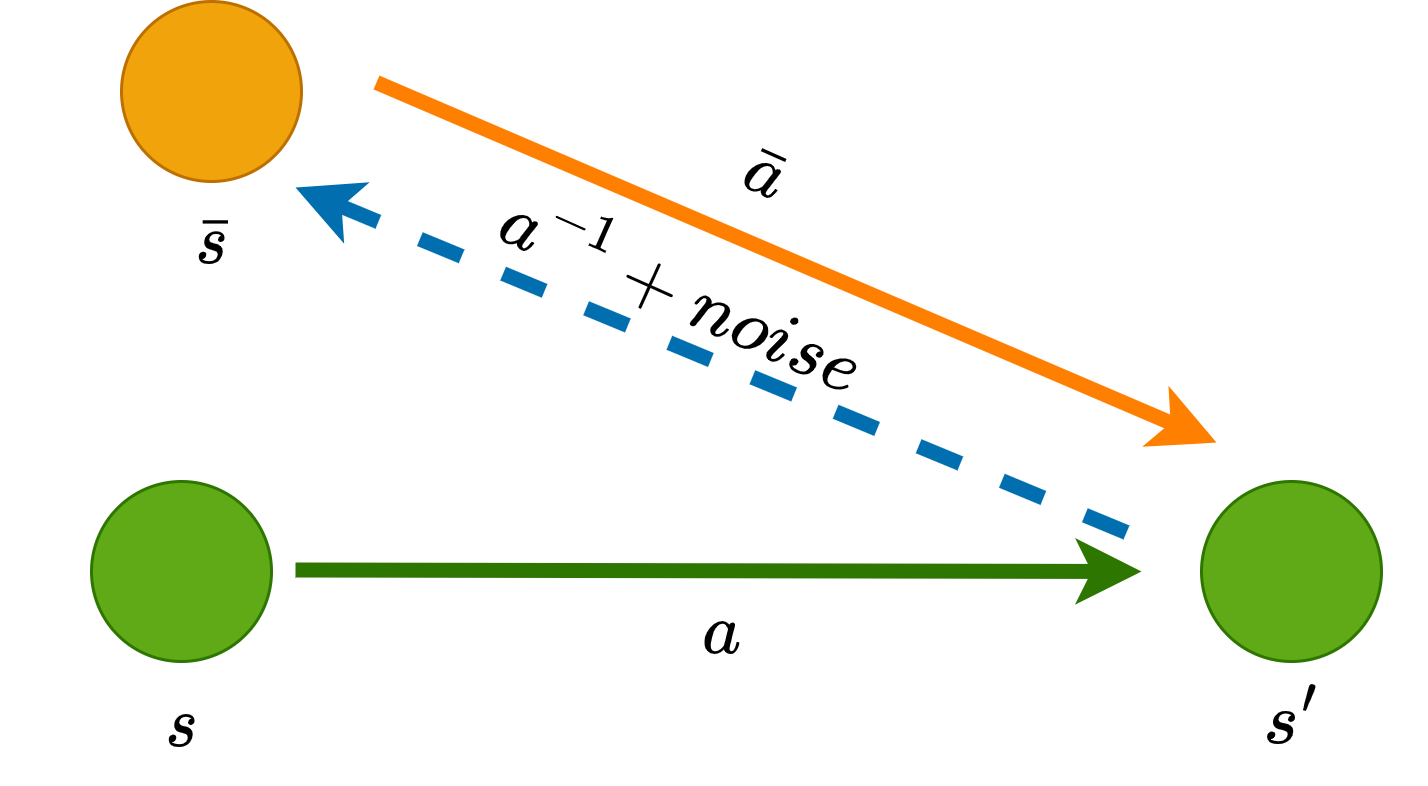}}\\
\centering
\subfloat[Breadth and depth simulation]{\includegraphics[width=0.65\linewidth]{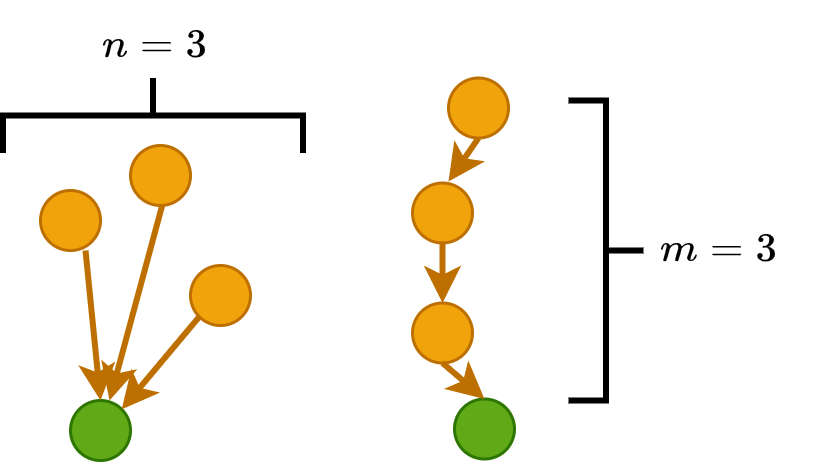}\label{fig:ts_type}}
\caption{\textbf{Concept of transition simulation.} (a) shows how simulated transitions (orange) are distributed around expert trajectories (green). 
(b) shows how transition simulation is accomplished for one specific transition, which is being described in Algorithm~\ref{alg:ts}. (c) shows the breadth and depth simulation of our method for $n=3$ or $m=3$.}
\label{fig:abs_ts}
\end{figure}

\begin{algorithm}[t]
\caption{Transition Simulation}
\textbf{Input}: N steps of expert transitions $\{T_1, T_2, ..., T_N\}$, buffer $B$, valid checking function $\mathrm{transition\_valid}$, parameter $n$ and $m$.

\textbf{Output}: Observation-action pair Buffer $B$.
\begin{algorithmic}[1]
\For{$i=1\dots N$}
\State $B.\mathrm{append}(\Pi(T_i.s), T_i.a)$
\For{$j= 1\dots n$} \# breadth expansion
\State $T = T_i$
\For{$k=1 \dots m$} \# depth expansion
\State $(s, a, s') = T$
\State Create $T^{sim}$ using Equation \ref{eqn:a_hat_inv}, \ref{eqn:s_hat}, \ref{eqn:sim_transition}.
\If {$\mathrm{transition\_valid}(T^{sim})$}
\State $B.\mathrm{append}(\Pi(T^{sim}.s), T^{sim}.a)$ 
\Else{ \# data balancing}
\State $B.\mathrm{append}(\Pi(T.s), T.a)$ 
\EndIf
\State $T=((C, \bar{p}+\bar{a}^{-1}, r), \bar{a}, \bar{s})$
\EndFor
\EndFor
\EndFor

\end{algorithmic}
\label{alg:ts}
\end{algorithm}

Few-shot imitation learning often suffers from poor generalization due to insufficient demonstration data, which is not enough to cover all the variations in the workspace. As a result, agents are usually stuck in situations with unseen and out-of-distribution observations. To solve this problem, we propose Transition Simulation that can create simulated expert transitions $T^{sim}$ from existing pre-collected expert transition $T$, as shown in Figure~\ref{fig:abs_ts}.

Our key idea is to simulate new expert transitions by creating new observations using the projection function $\Pi$ and the rigid transformation of the robot gripper. Given a fixed point cloud $C$, one can generate a depth image at an arbitrary camera position. This means we can simulate the observation $I$ as if the gripper is located at an arbitrary pose in the workspace (a similar method has been used by~\cite{song_grasping_2020} to serve as a forward model for rendering action views). We choose to simulate new transitions that are nearby to the demonstrated trajectories; these simulated transitions serve as a funnel that guides the agent toward expert states. Given an expert transition $T=(s, a, s')$ where $s=(C, p, r), a=(\lambda, \Delta p), s'=(C', p', r')$, we first define the inverse of $a$ as inverting the delta pose $\Delta p$: $a^{-1} = (\lambda, -\Delta p)$. $\bar{a}^{-1}$ is defined as adding a Gaussian noise to $a^{-1}$: 
\begin{equation}
    \bar{a}^{-1} = a^{-1} + \epsilon,\quad \epsilon \sim \mathcal{N}(0, \sigma),
    \label{eqn:a_hat_inv}
\end{equation}
where $\sigma$ is a hyper-parameter for the standard deviation of the Gaussian noise. Then we create a simulated position $\bar{p} = p' + \bar{a}^{-1}$ by applying the delta action in $\bar{a}^{-1}$ to $p'$. Since $\bar{a}^{-1}$ is close to $a^{-1}$, $\bar{p}$ is within the vicinity of $p$. Thus a simulated state $\bar{s}$ can be created by replacing $p$ with $\bar{p}$ in $s$:
\begin{equation}
    \bar{s} = (C, \bar{p}, r).
    \label{eqn:s_hat}
\end{equation}
Next, we can take the inverse of $\bar{a}^{-1}$, $\bar{a}$, and form a simulated expert transition: 
\begin{equation}
    T^{sim}=(\bar{s}, \bar{a}, s').
    \label{eqn:sim_transition}
\end{equation}
In the end, we can create a new expert observation-action pair by querying $(\Pi(C, \bar{p}), \bar{a})$. After this process, we can further simulate a state prior to $\bar{s}$ by applying $\bar{a}^{-1}$ to $\bar{p}$ again: $(C, \bar{p}+\bar{a}^{-1}, r)$.

Intuitively, our method views an expert trajectory as a tree with a single path where the initial state is a leaf and the terminal state is the root. Our method creates new expert transitions by expanding the tree with new branches within the vicinity of the existing nodes in the tree. In contrast, standard data augmentation techniques (e.g., applying a rotation or translation to the state and action pair~\cite{zeng2020transporter}) can only duplicate an expert trajectory path under a fixed transformation applied to all nodes and edges, rather than creating new connected paths. Moreover, our method can be used in conjunction with the standard data augmentation to further improve the sample efficiency. 

The full algorithm is shown in Algorithm~\ref{alg:ts}. Notice that $n$ and $m$ are hyper-parameters that control the number of breadths and the number of depth for the simulated transitions, illustrated in Figure~\ref{fig:ts_type}. Line 8-11 is a data balancing (DB) technique such that when the simulated transition is not valid, a duplication of the original transition will be generated (we empirically find that this data balancing will improve the performance, see Section~\ref{sec:ablation}). In our work, the $\mathrm{transition\_valid}$ function is defined as that the simulated position $\bar{p}$ is contact-free (i.e., $\bar{p}\in \mathcal{T}$) and the gripper is not holding any object. However, the method can be extended to placing transitions while holding objects to enhance placing behaviors for pick and place tasks. Line 13 creates a transition prior to $\bar{s}$ for depth expansion.

\begin{figure*}[t]
\centering
\newlength{\env}
\setlength{\env}{0.115\linewidth}
\subfloat[Block Stacking]{
\includegraphics[width=\env]{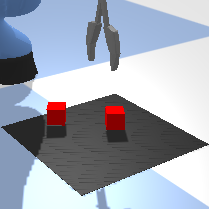}
\includegraphics[width=\env]{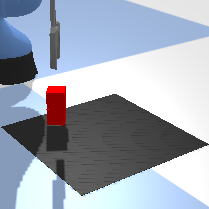}
}
\subfloat[Block in Bowl]{
\includegraphics[width=\env]{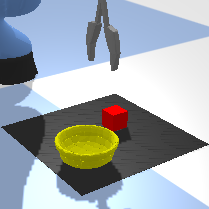}
\includegraphics[width=\env]{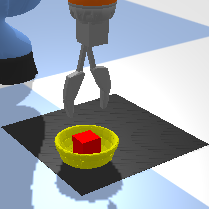}
}
\subfloat[Drawer Opening]{
\includegraphics[width=\env]{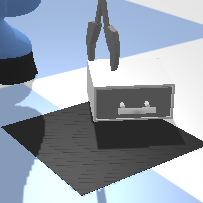}
\includegraphics[width=\env]{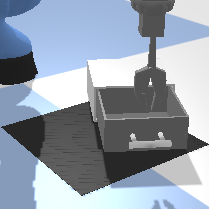}
}
\subfloat[Shoe Packing]{
\includegraphics[width=\env]{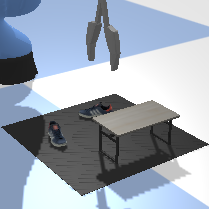}
\includegraphics[width=\env]{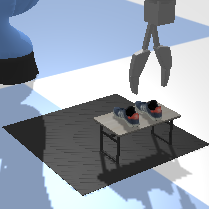}
}\\
\subfloat[Table Tidying]{
\includegraphics[width=\env]{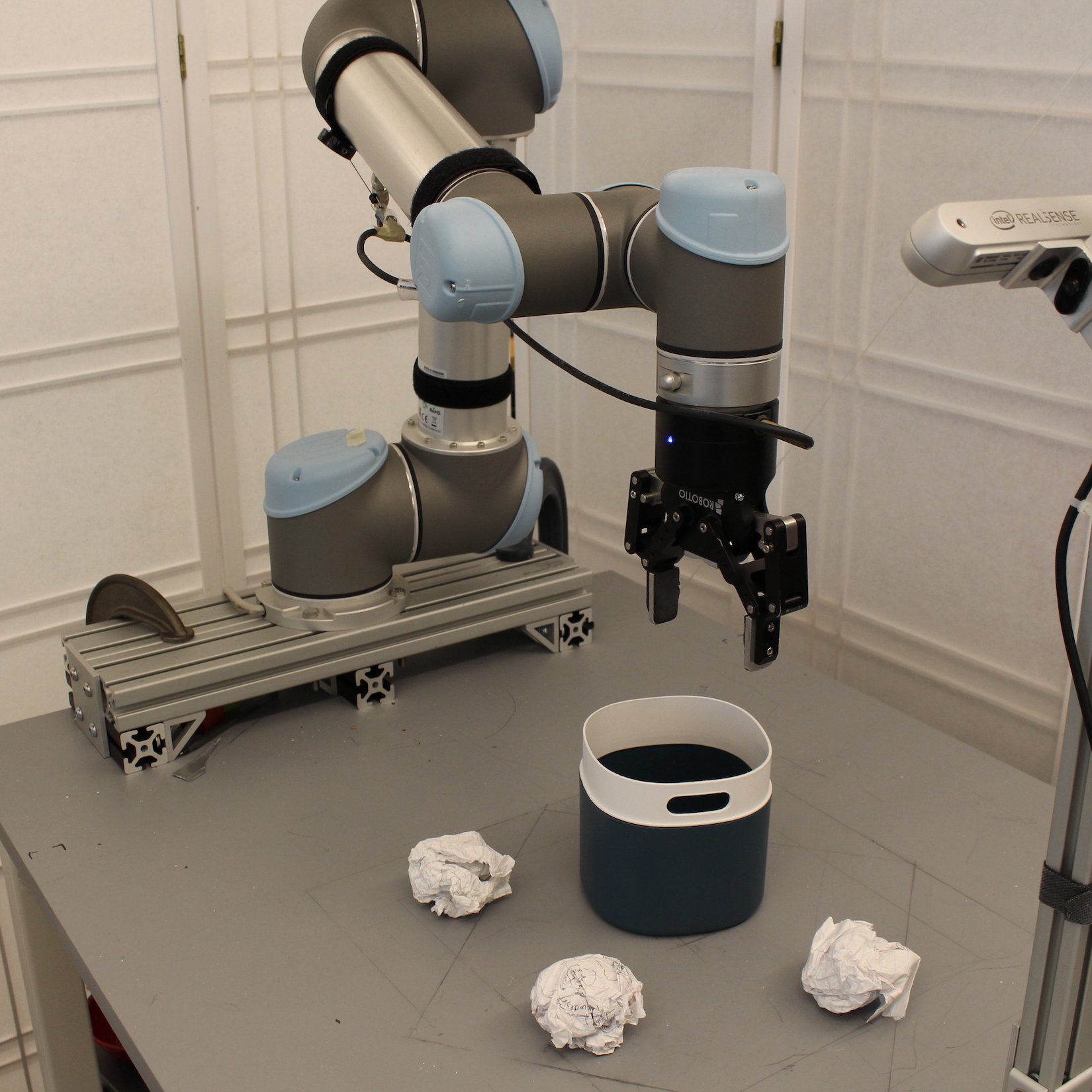}
\includegraphics[width=\env]{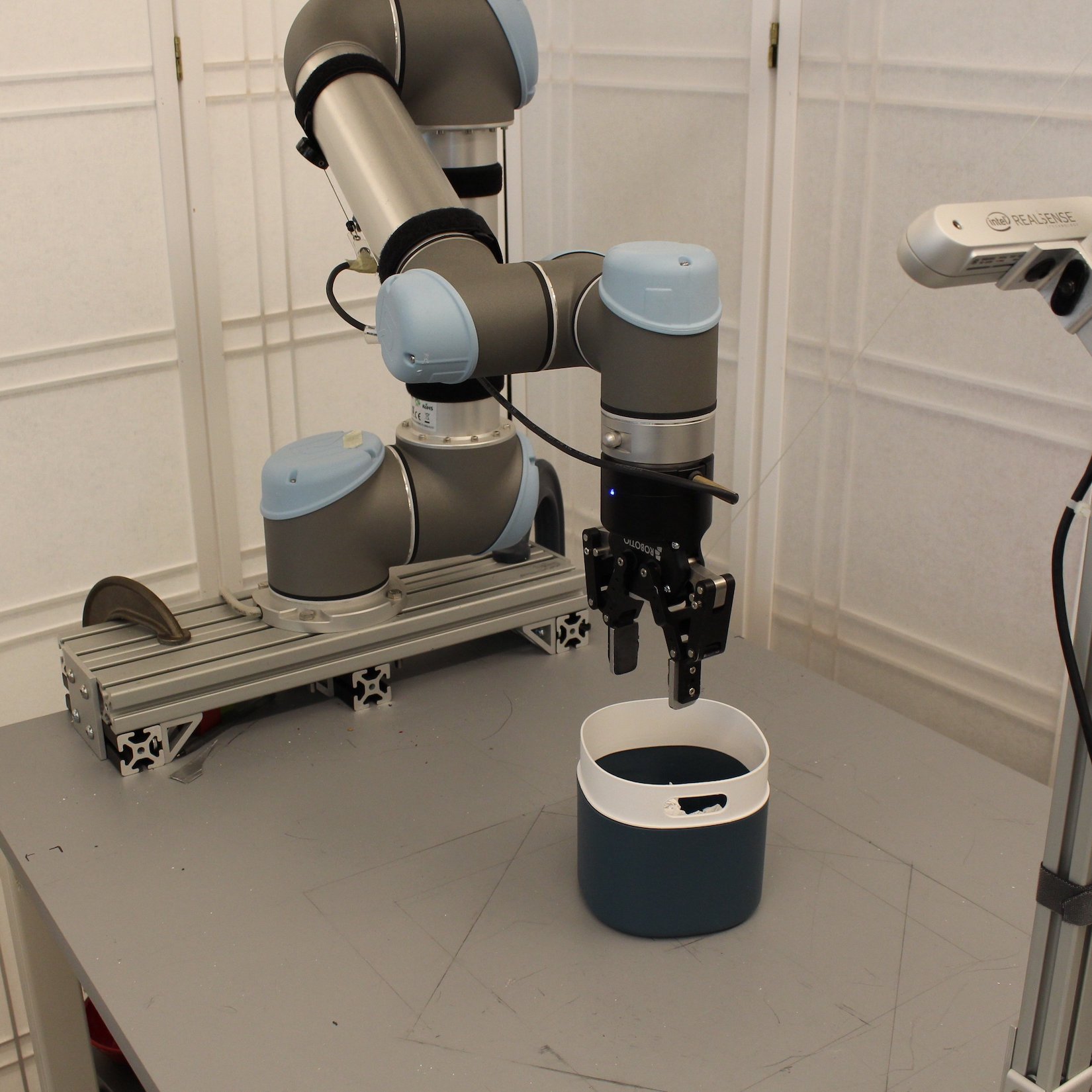}
}
\subfloat[Block in Bowl]{
\includegraphics[width=\env]{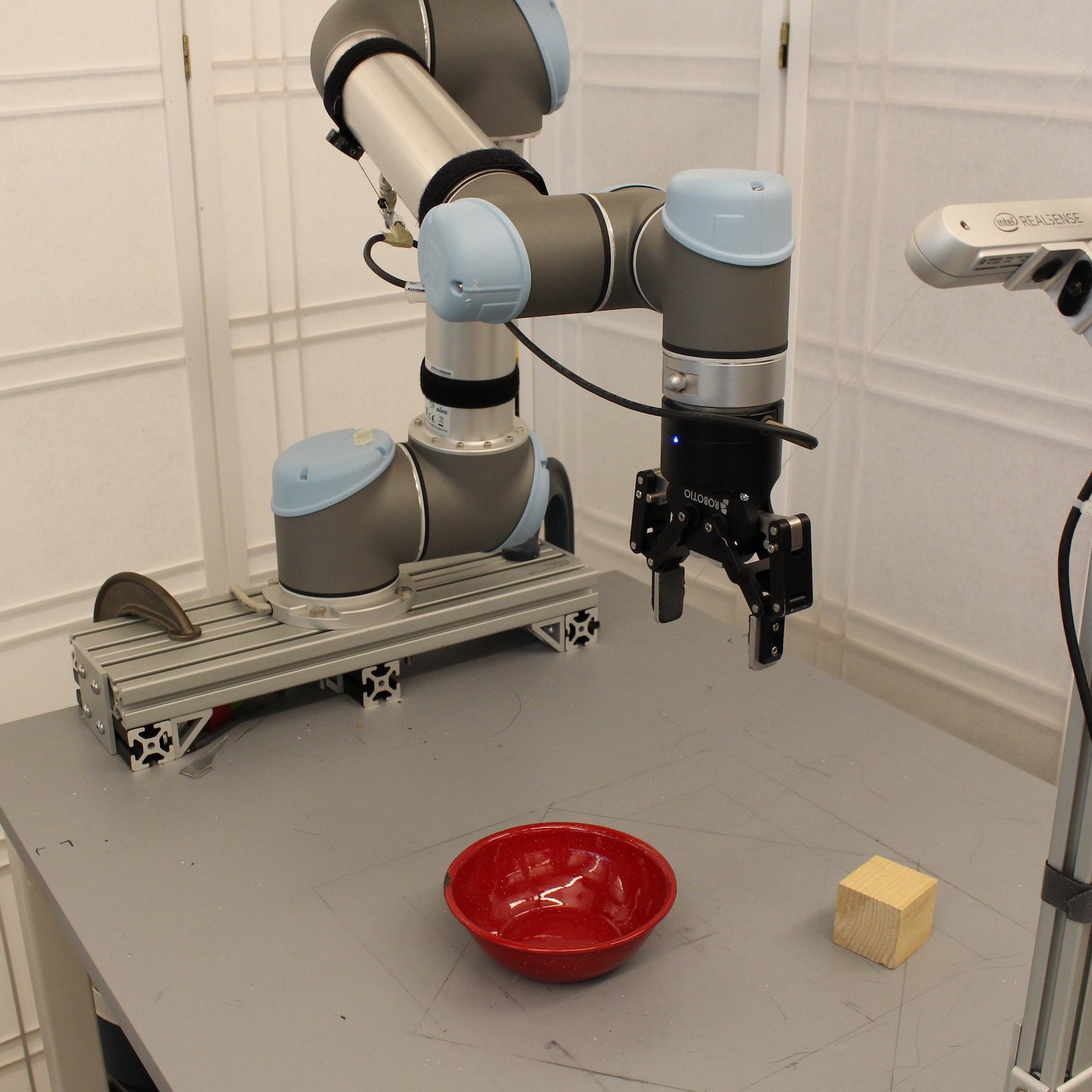}
\includegraphics[width=\env]{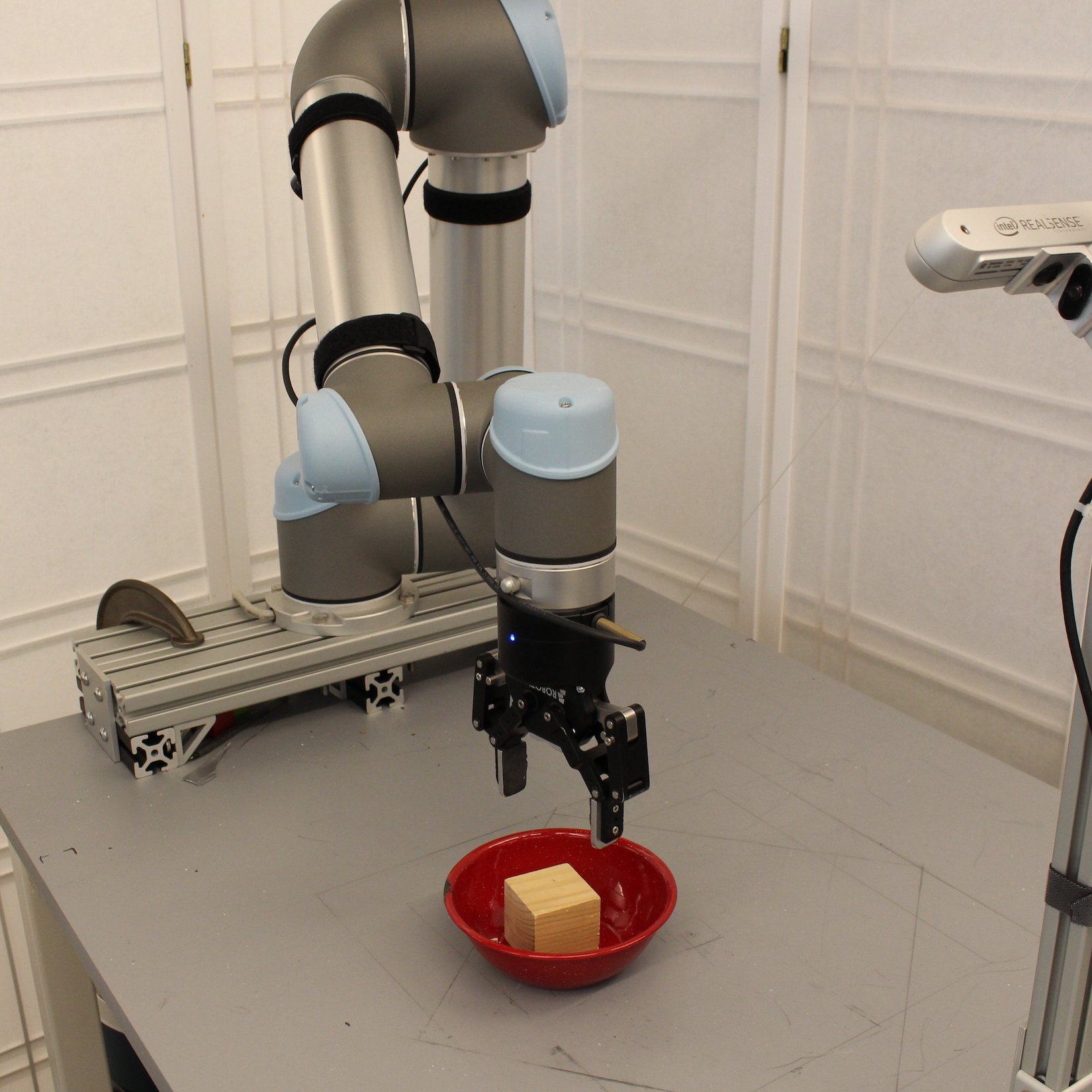}
}
\subfloat[Drawer Opening]{
\includegraphics[width=\env]{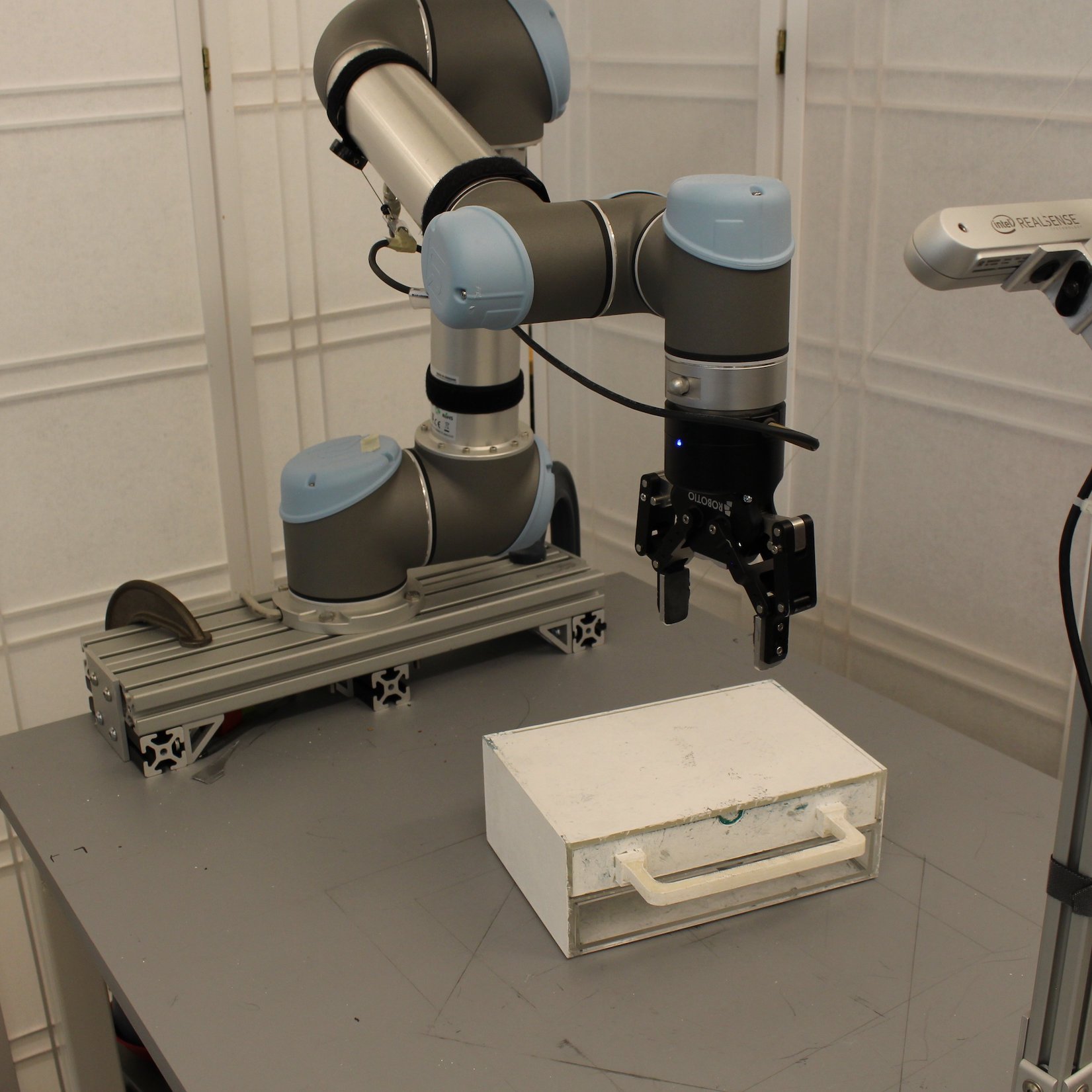}
\includegraphics[width=\env]{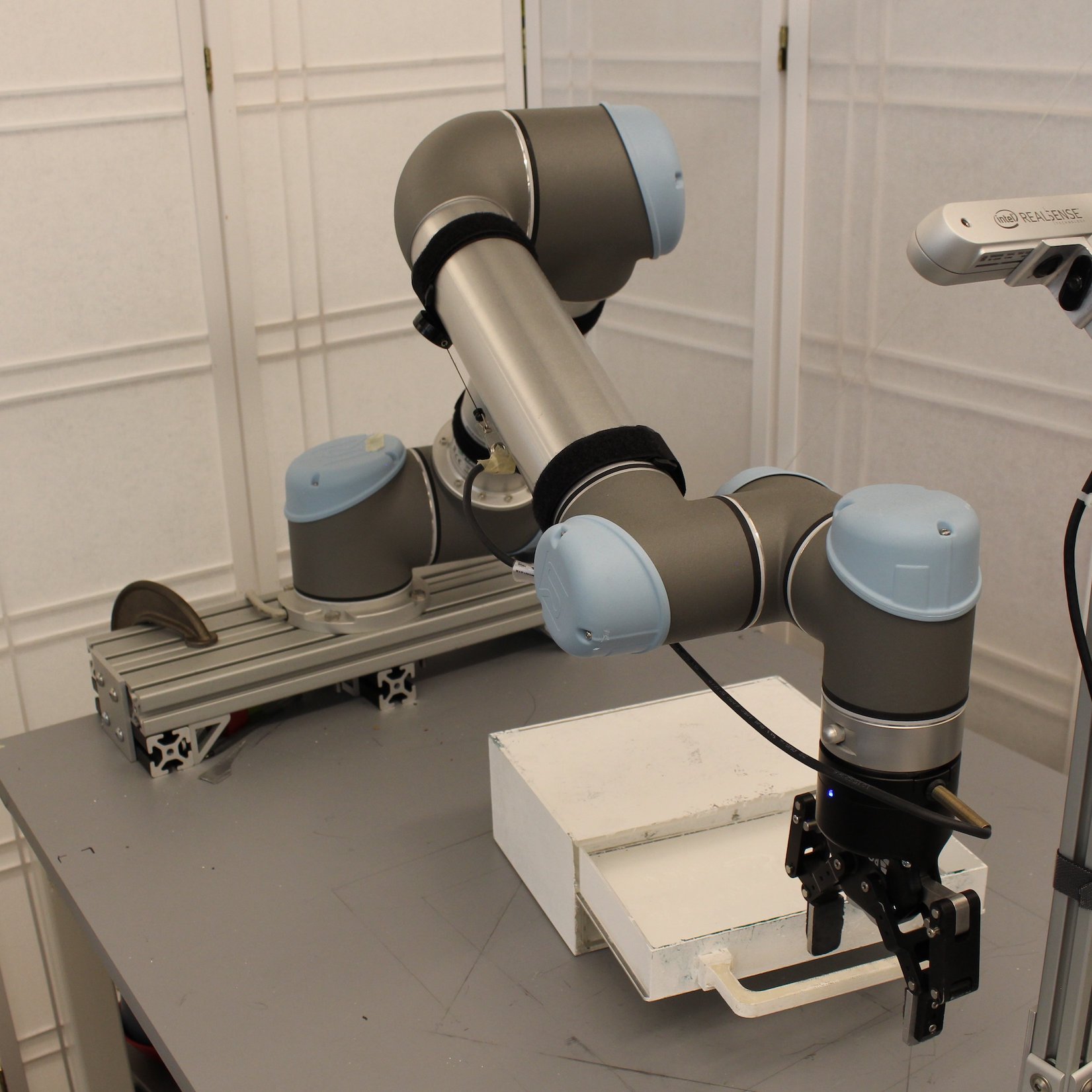}
}
\subfloat[Shoe Packing]{
\includegraphics[width=\env]{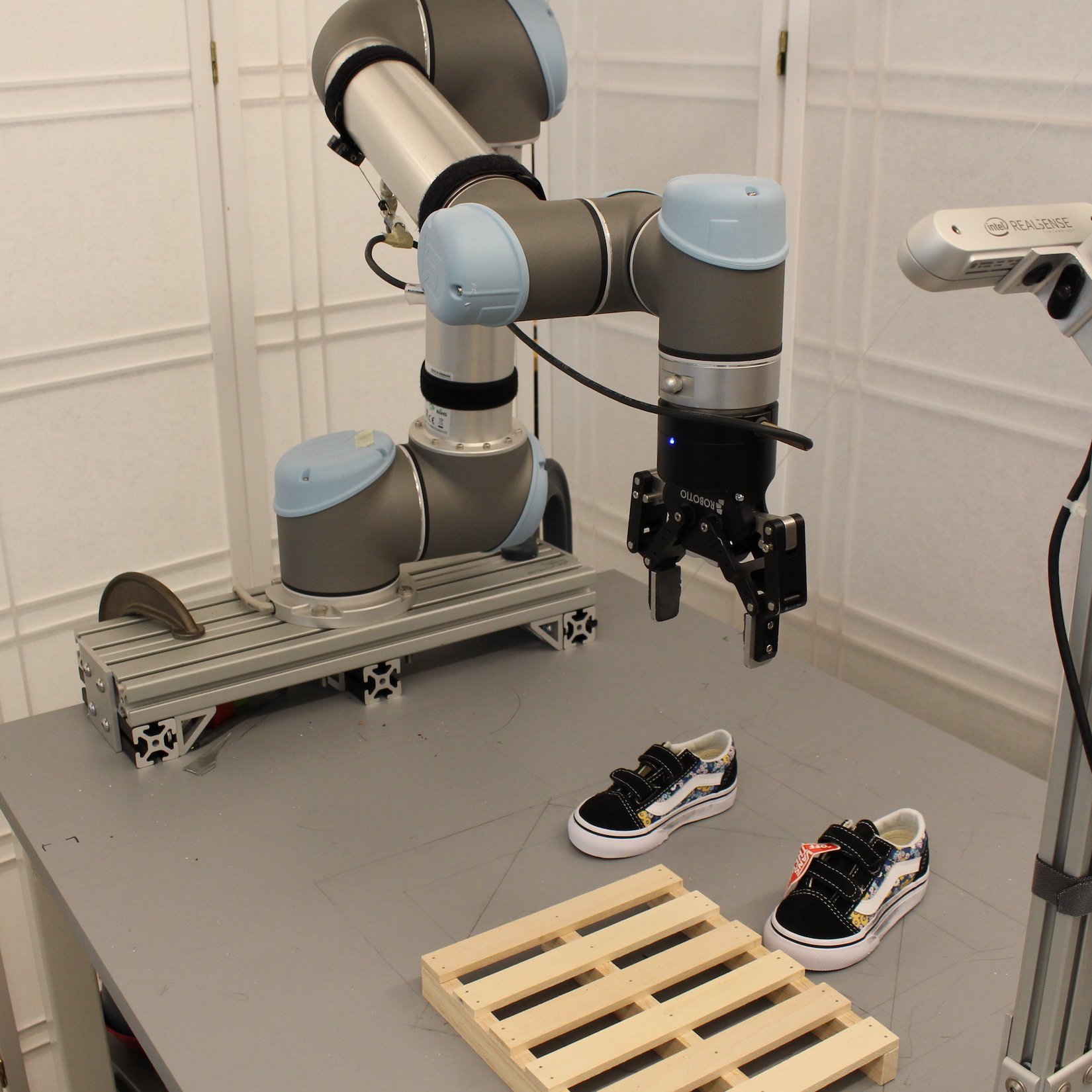}
\includegraphics[width=\env]{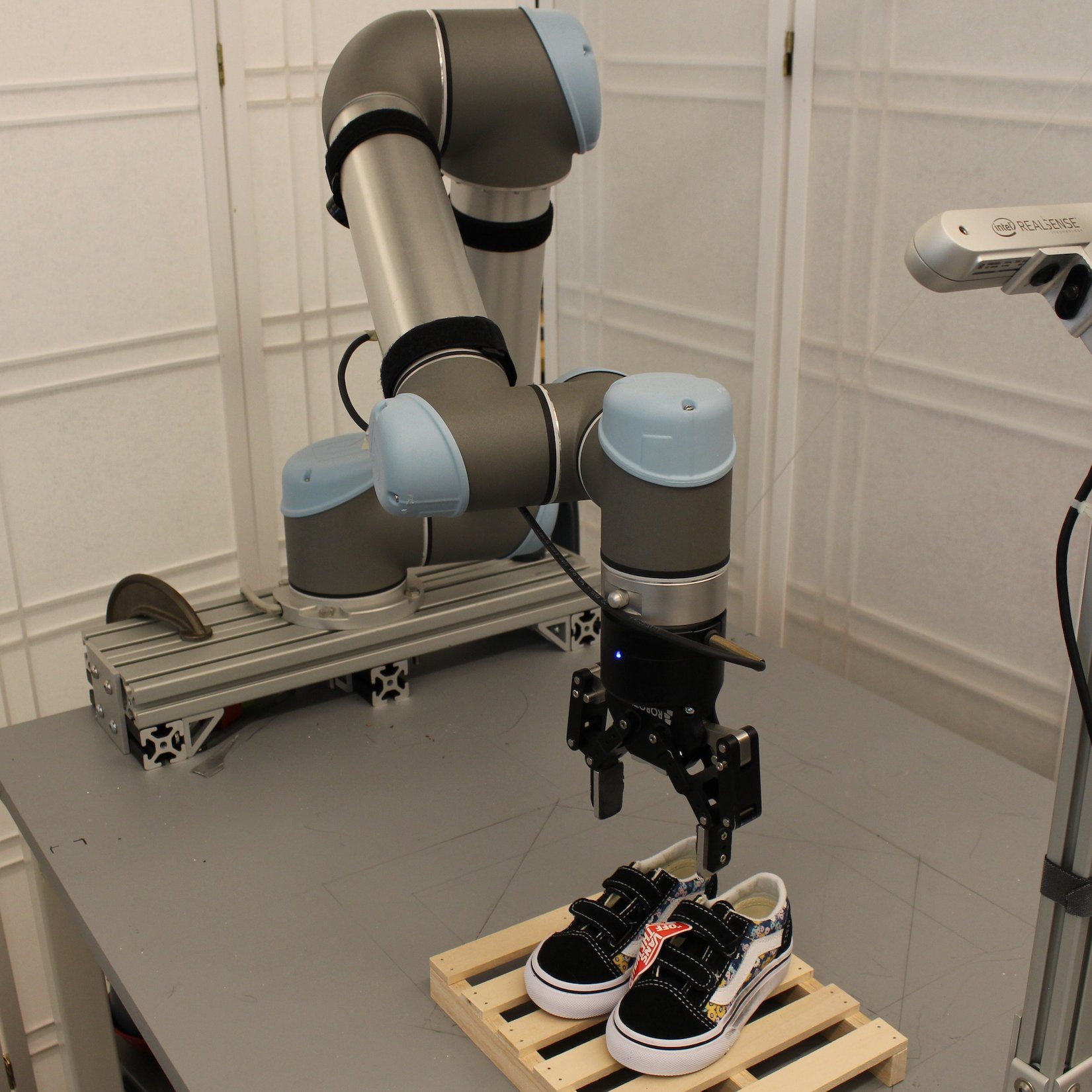}
}
\caption{\textbf{Environments and goal states.} (a)-(d) Simulation environments in PyBullet~\cite{pybullet}; (e)-(h) Real-world environments on a physical robot.}
\label{fig:sim_envs}
\end{figure*}

\subsection{Equivariant Behavioral Cloning (Equi BC)} 
\label{sec:equi_policy}
The second main component in our method is an equivariant model that improves sample efficiency through enforcing $\OO(2)$ (i.e., the group of all planar rotation and reflection) symmetry in the network architecture. Inspired by prior work~\cite{so2rl,wang2022onrobot} utilizing an equivariant actor function in an actor-critic algorithm, we apply a similar approach to behavior cloning. Specifically, the expert policy function $\pi^*$ can be written as a $\OO(2)$-equivariant function:
\begin{equation}
    \pi^*(gs) = g\pi^*(s),
\end{equation}
where $g\in \OO(2)$. That is to say, when the manipulation scene is rotated and/or reflected by $g$, the action should be rotated and/or reflected accordingly. $g$ acts on the state $s$ through rotating and reflecting the point cloud $C$ and the pose of the gripper $p$ in the state; $g$ acts on the action $a$ through rotating and reflecting the $(x, y)$ vector in $\Delta p$ and reflecting the $\theta$ component in $\Delta p$. Formally, we define
\begin{equation}
    gs=(gC, gp, r); ga=(\lambda, (R_g(\Delta x, \Delta y), \Delta z, f\Delta \theta)),
\end{equation}
where $R_g$ is the $2\times 2$ transformation matrix with respect to $g$, and $f \in \{1, -1\}$ is the reflection component in $g$. Notice that our observation projection function $\Pi: s\mapsto I$ is an equivariant function, i.e., $\Pi(gs) = g\Pi(s) = gI$ because the image $I$ encodes both the point cloud $C$ and the gripper pose $p$. We can then define our policy network $\pi: I\mapsto a$ as an equivariant function:
\begin{equation}
    \pi(gI) = g\pi(I).
\end{equation}

We implement such an equivariant policy network using Steerable CNNs~\cite{steerable_cnns,e2cnn}. Steerable CNNs share weights across different group elements, enabling the network to automatically generalize to all group transformations and, consequently, improve the sample efficiency. We use a similar network architecture as the prior work~\cite{so2rl}. For a more comprehensive introduction on how to implement such an equivariant network, please refer to~\cite{so2rl}.


\section{EXPERIMENTS}

\begin{table*}[t]
  \setlength\tabcolsep{2.3pt}
  \centering
    \caption{Baseline comparisons in simulation. Task success rates averaged over four runs.}
  \begin{tabular}{@{}lcccccccccccccccccccc@{}}
  \toprule
  & \multicolumn{4}{c}{Block Stacking} & \multicolumn{4}{c}{Block in Bowl} & \multicolumn{4}{c}{Drawer Opening} & \multicolumn{4}{c}{Shoe Packing} & \\
  \cmidrule(lr){2-5} \cmidrule(lr){6-9} \cmidrule(lr){10-13} \cmidrule(lr){14-17} 
  Method & 1 & 5 & 10 & 100 & 1 & 5 & 10 & 100 & 1 & 5 & 10 & 100 & 1 & 5 & 10 & 100  \\
  \midrule
  CNN BC                           & 18.5 & 73.5 & 79.0 & 90.5 & 19.0 & 93.5 & 99.0 & \textbf{100} & 31.0 & 66.5 & 76.0 & 88.5 & 2.0 & 4.7 & 12.0 & 20.0 \\
  Implicit BC         & 11.0 & 9.5 & 51.0 & 80.5 &  13.0 & \textbf{99.5} & \textbf{100} & \textbf{100} & 31.0 & 63.5 & 71.5 & 81.5 & 0.5 & 5.5 & 12.0 & 13.0 \\
CNN BC + TS                       & 41.0 & 75.0 & 87.0 & 92.0 & 52.2 & 91.0 & 96.5 & 98.5 & 53.5 & 76.0 & 75.0 & 84.5 & 7.5 & 13.0 & 22.0 & 26.0 \\
  \midrule
  \textbf{Equi BC (Ours)} & 33.5 & 87.5 & 93.0 & \textbf{100} & 46.5 & \textbf{99.5} & 99.5 & \textbf{100} & 62.5 & 88.5 & 91.0 & \textbf{100} & 1.5 & 22.5 & 39.5 & \textbf{75.0}\\
  \textbf{SEIL (Ours)}      & \textbf{71.5} & \textbf{99.5} & \textbf{98.5} & \textbf{100} & \textbf{75.0} & 98.0 & \textbf{100} & \textbf{100} & \textbf{78.5} & 87.5 & \textbf{93.5} & 96.5 & \textbf{16.5} & \textbf{57.3} & \textbf{68.0} & 72.0\\
  \bottomrule
  \end{tabular}
  \label{table:sim_result}
\end{table*}

\begin{figure*}[t]
\includegraphics[width=\textwidth]{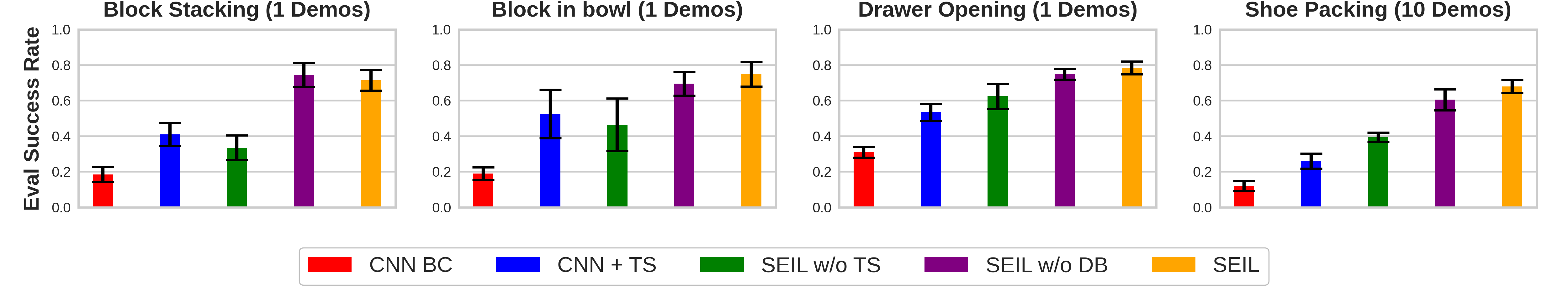}
\caption{\textbf{Ablation study.} The figure shows how every component of our method improves performance. Results averaged over four runs. }
\label{fig:ablation}
\end{figure*}

\begin{figure}[t]
\includegraphics[width=\linewidth]{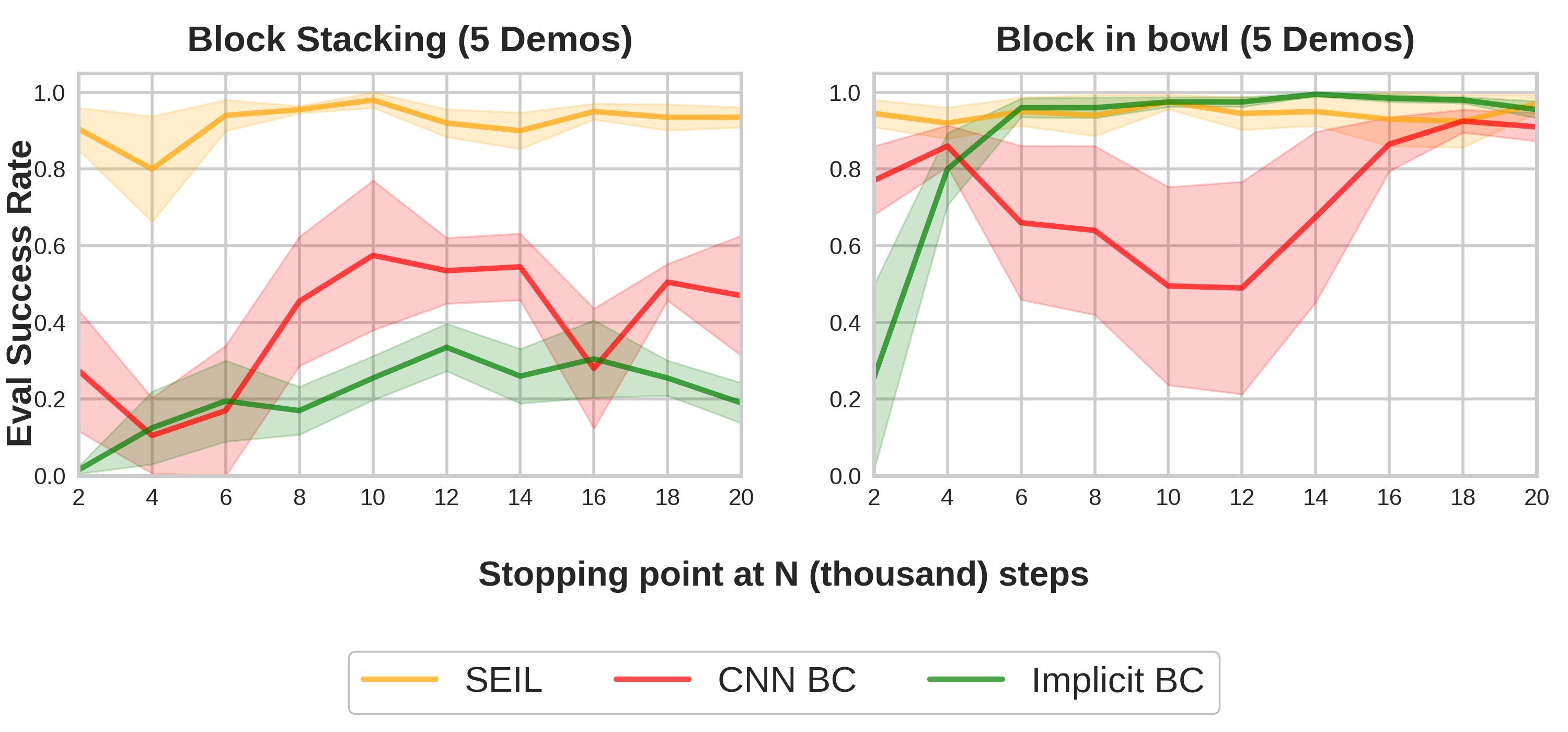}
\caption{\textbf{Convergence.} The plot shows the performance of the learned policy at different numbers of training steps. Results averaged over four runs. Shading represents standard error. }
\label{fig:conver}
\end{figure}


\subsection{Simulation experiments}
\label{sec:exp_sim}
\subsubsection{Tasks}
In this experiment, we evaluate SEIL in four tasks in simulation using PyBullet~\cite{pybullet}: Block Stacking, Block in Bowl, Drawer Opening, and Shoe Packing (Figure~\ref{fig:sim_envs})\footnote{The first three environments are from the BulletArm benchmark~\cite{bulletarm}.}. For all environments, the maximum moving distance per step are limited as $\Delta x, \Delta y, \Delta z \in [-2cm, 2cm]$; $\Delta \theta \in [-\frac{\pi}{8}, \frac{\pi}{8}]$; $\lambda\in [0, 1]$ where $0$ indicates fully close and $1$ indicates fully open. We use a state-based planner to gather $N$ episodes of expert demonstrations. For Transition Simulation, the breadth expansion parameter $n$ is set to 4 and the depth expansion parameter $m$ is set to 1. The Gaussian noise standard deviation $\sigma$ is 0.4. In addition to TS, we also apply a rotational data augmentation where each expert observation-action pair (including the ones created using TS) is augmented 64 times using random $\SO(2)$ rotations. The models are trained with MSE loss for 20,000 training steps. For each run, we evaluate the model for 50 episodes every 2000 training steps (eight evaluations in total). We take the highest success rate of all eight evaluations as the result of one run. Final results are reported by taking the average over four runs.

\subsubsection{Baseline Comparison}

We compare SEIL with the following baselines: 1) CNN BC: standard Behavior Cloning method implemented using conventional CNNs; 2) CNN BC + TS: 1) equipped with our Transition Simulation; 3) implicit BC~\cite{ibc}: an energy-based model (EBM) that takes observation and actions as input, outputting corresponding scalar energies (implicit BC uses infoNCE loss instead of MSE loss); 4) Equi BC: Behavior Cloning implemented using our equivariant policy network.
We run each method in each environment with $N=1, 5, 10, 100$ demonstration episodes.

Table~\ref{table:sim_result} shows the results. First, notice that our SEIL algorithm has the best overall performance. It allows the agent to directly generalize to unseen scenarios with a few expert demonstrations. In particular, given only one episode of expert data, SEIL shows a large performance gap compared with the baselines. When the agent is provided with more data, our method still outperforms the baselines with a high success rate. Second, comparing CNN BC + TS and Equi BC with CNN BC, both TS and the equivariant policy function generally improve the performance of CNN BC, especially with fewer demonstrations. This indicates that both TS and the equivariant policy can be used as a plug-in in a standard BC method to boost performance. 
Another interesting finding is that equivariant models converge faster and is more stable than conventional CNN models due to data efficiency. Figure~\ref{fig:conver} shows the evaluation curve of SEIL compared with CNN BC and implicit BC in two environments. CNN models converge slower and the performance is not stable at different training steps. Consequently, frequent evaluations in different training steps are necessary for CNN models to find the best performance. In contrast, Equivariant models are able to converge within 2k training steps and keep their robustness during eight evaluations, which is very useful for on-robot learning where frequent evaluation is time-consuming.

\subsubsection{Ablation Study}
\label{sec:ablation}
In this experiment, we study the importance of each component of our method in an ablation study. We consider the following baselines: 1) CNN BC: a baseline without any of our contribution; 2) SEIL w/o Equi: our method using a standard CNN network instead of an equivariant network (i.e., only having the TS in Section~\ref{sec:ts}. This is equivalent to equipping CNN BC with TS); 3) SEIL w/o DB: our method without data balancing (line 8-11 in Algorithm~\ref{alg:ts}); 4) SEIL w/o TS: our method without Transition Simulation completely (i.e., only having the equivariant policy in Section~\ref{sec:equi_policy}); 5) SEIL: our full algorithm. Figure~\ref{fig:ablation} shows the result. First, notice that removing TS (green) or removing the equivariant policy (blue) both lead to a significant performance drop compared with our full algorithm (yellow), showing that they are equally important in our method. Meanwhile, both green and blue are significantly better than the baseline standard BC method (red). Second, comparing the purple and the yellow, removing the data balancing slightly impedes the performance of our method.

\subsection{Real-world experiments}

\begin{table*}[t]
  \setlength\tabcolsep{2.3pt}
  \centering
    \caption{Baseline comparisons in real-world experiments. Task success rates averaged over 30 episodes across three seeds.}
  \begin{tabular}{@{}lcccccccccccccccc@{}}
  \toprule
  & \multicolumn{3}{c}{Trash Tidying} & \multicolumn{3}{c}{Block in Bowl} & \multicolumn{3}{c}{Drawer Opening} & \multicolumn{3}{c}{Shoe Packing} & \\
  \cmidrule(lr){2-4} \cmidrule(lr){5-7} \cmidrule(lr){8-10} \cmidrule(lr){11-13} 
  Method & 1 & 5 & 10 & 1 & 5 & 10 & 1 & 5 & 10 & 1 & 5 & 10  \\
  \midrule
  
  CNN BC                        & 0.0 & 3.3 & 20.0 & 0.0 & 50.0 & 76.7 & 23.3 & 60.0 & 70.0 & 10.0 & 46.7 & 70.0 \\
  \textbf{SEIL (Ours)}               & \textbf{30.0} & \textbf{60.0} & \textbf{93.3} & \textbf{36.7} & \textbf{90.0} & \textbf{100.0} & \textbf{86.7} & \textbf{96.7} & \textbf{100.0} & \textbf{13.3} & \textbf{60.0} & \textbf{90.0}   \\
  \bottomrule
  \end{tabular}
  \label{table:real_results}
\end{table*}

  

\begin{figure}[t]
\centering
\includegraphics[width=0.8\linewidth]{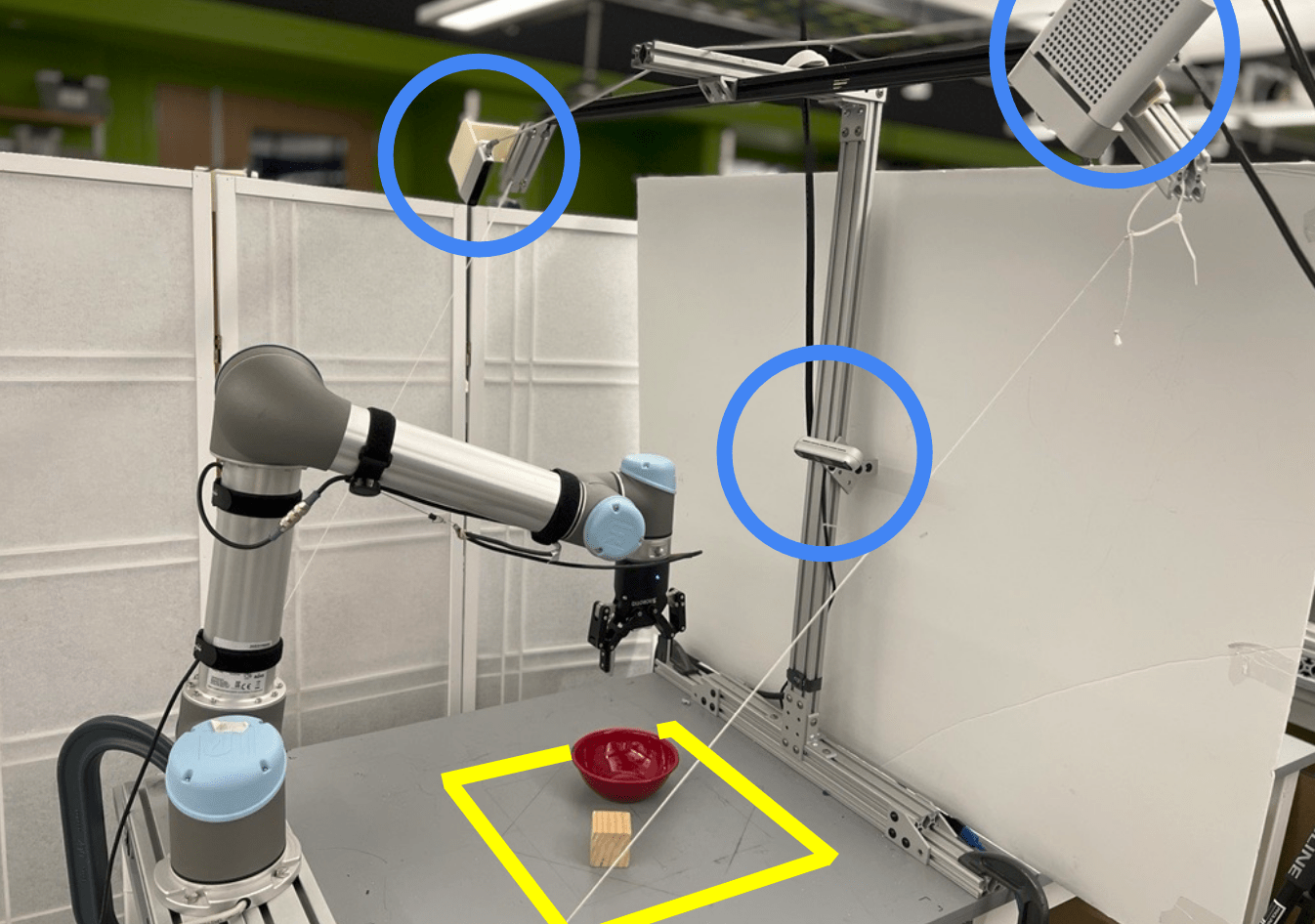}
\caption{\textbf{Real-world environment and settings.} Three cameras are marked in blue circle. The workspace is highlighted in yellow.}
\label{fig:real_setting}
\end{figure}


\subsubsection{System setup}
Our real-world experimental platform consists of a Universal Robots UR5 robotic arm equipped with a Robotiq 2F-85 parallel-jaw gripper and three depth cameras (one Intel Realsense D455, one Occipital structure sensor, and one Microsoft Azure Kinect DK), as is shown in Figure~\ref{fig:real_setting}. The point clouds from three cameras are fused into the point cloud $C$ in our state space. Our workstation has an Intel Core i7-9700k CPU (3.60GHz) and an Nvidia RTX 2080Ti GPU. We experiment with four tasks in the real world: Table Tidying, Block in Bowl, Drawer Opening, and shoe Packing (Figure~\ref{fig:sim_envs} e-f). The expert demonstrations are provided by a person using a PS4 controller. In each task, we train the model for 20k training steps and evaluate the success rate over 10 episodes. In Table~\ref{table:real_results}, we report the performance averaged over three random seeds (30 episodes in total). Other experimental parameters match the simulation experiment in Section~\ref{sec:exp_sim}.

\subsubsection{Results}
\begin{figure}[t]
\centering
\subfloat[CNN pick]{\includegraphics[width=0.235\linewidth]{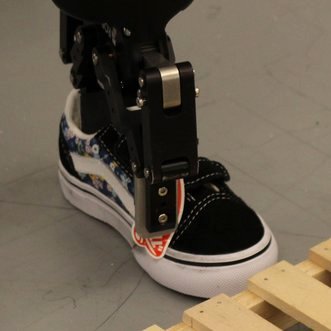}
}
\subfloat[CNN place]{\includegraphics[width=0.235\linewidth]{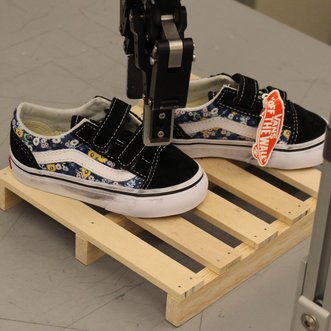}
}
\subfloat[SEIL pick]{\includegraphics[width=0.235\linewidth]{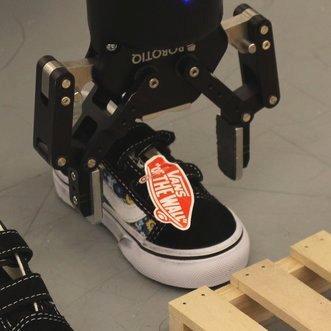}
}
\subfloat[SEIL place]{\includegraphics[width=0.235\linewidth]{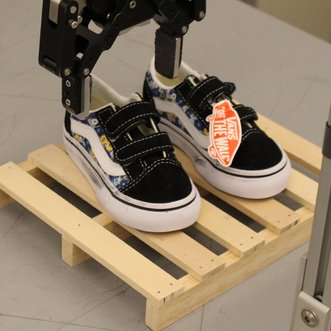}
}
\caption{\textbf{Comparison of SEIL and CNN BC in Shoe Packing.} SEIL is able to pack the shoes with reasonable orientations, but CNN cannot.}
\label{fig:failure}
\end{figure}

As shown in Table~\ref{table:real_results}, in the real world experiments, our SEIL constantly outperforms the baseline CNN BC, especially when there are fewer expert demonstrations. In Trash Tidying, SEIL achieves 93.3\% success rate with ten expert demonstrations, whereas CNN BC can only reach 20\%. The common failure mode for CNN BC is that it fails to learn a policy to pick up any trash. Our SEIL algorithm, on the other hand, generalizes well from the limited demonstrations. In Block in Bowl with ten demos, SEIL succeeds in all 30 evaluations, whereas CNN often fails to accurately pick up the block (e.g., in 2 of the failures, it picks up the bowl instead of the block) and only achieves 76.7\% success rate. With only five demons, SEIL can still achieve a 90\% success rate whereas CNN BC can only reach 50\%. In Drawer Opening with ten demos, SEIL reaches 100\% success rate. CNN BC often fails to stick the finger into the gap between the drawer and the handle (in all failures), and only reaches 70\% success rate. In Shoe Packing with ten demons, SEIL not only outperforms the baseline by 20\%, but also packs the shoes with the proper orientations (Figure~\ref{fig:failure}). 

\section{CONCLUSIONS AND LIMITATIONS}
This paper proposes Transition Simulation, a novel trajectory generation algorithm that can augment the existing expert demonstration data. This paper further combines Transition Simulation with equivariant policy learning and introduces SEIL, a sample-efficient imitation learning algorithm that can solve robotic manipulation tasks within ten or fewer expert demonstrations. The major limitation of our work is that Transition Simulation can only simulate transitions without contact between the gripper and the environment. This can be improved by learning an environment model to predict the outcome of the transition with contact. Another limitation is that our observations are based on depth images without color information, preventing the agent from solving more complex tasks like color-based sorting. In future work, we plan to include RGB channels in the point cloud and in the projection image. Another direction for future work is to extend our Transition Simulation to, for example, simulate on-policy transitions for reinforcement learning.

\bibliographystyle{IEEEtran}
\bibliography{references}

\end{document}